\documentclass[conference]{IEEEtran}
\makeatletter\if@twocolumn\PassOptionsToPackage{switch}{lineno}\else\fi\makeatother

\usepackage{graphicx}
\usepackage[T1]{fontenc}
\ifCLASSINFOpdf
\else
\fi
\usepackage{amsmath}
\usepackage[cmintegrals]{newtxmath}
\usepackage{url,multirow,morefloats,floatflt,cancel,tfrupee}
\makeatletter

\AtBeginDocument{\@ifpackageloaded{textcomp}{}{\usepackage{textcomp}}}
\makeatother
\usepackage{colortbl}
\usepackage{xcolor}
\usepackage{pifont}
\usepackage[nointegrals]{wasysym}
\makeatletter

\def\mcWidth#1{\csname TY@F#1\endcsname+\tabcolsep}

\def\cAlignHack{\rightskip\@flushglue\leftskip\@flushglue\parindent\z@\parfillskip\z@skip}
\def\rAlignHack{\rightskip\z@skip\leftskip\@flushglue \parindent\z@\parfillskip\z@skip}

\@ifundefined{etal}{}{}

\usepackage{ifxetex}
\ifxetex\else\if@twocolumn\@ifpackageloaded{stfloats}{}{\usepackage{dblfloatfix}}\fi\fi

\AtBeginDocument{
\expandafter\ifx\csname eqalign\endcsname\relax
\def\eqalign#1{\null\vcenter{\def\\{\cr}\openup\jot\m@th
  \ialign{\strut$\displaystyle{##}$\hfil&$\displaystyle{{}##}$\hfil
      \crcr#1\crcr}}\,}
\fi
}

\AtBeginDocument{%
  \@ifpackageloaded{endfloat}%
   {\renewcommand\efloat@iwrite[1]{\immediate\expandafter\protected@write\csname efloat@post#1\endcsname{}}}{\newif\ifefloat@tables}%
}%

\def\BreakURLText#1{\@tfor\brk@tempa:=#1\do{\brk@tempa\hskip0pt}}
\let\lt=<
\let\gt=>
\def\processVert{\ifmmode|\else\textbar\fi}

\@ifundefined{subparagraph}{
\def\subparagraph{\@startsection{paragraph}{5}{2\parindent}{0ex plus 0.1ex minus 0.1ex}%
{0ex}{\normalfont\small\itshape}}%
}{}

\newcommand\role[1]{\unskip}
\newcommand\aucollab[1]{\unskip}
  
\@ifundefined{tsGraphicsScaleX}{\gdef\tsGraphicsScaleX{1}}{}
\@ifundefined{tsGraphicsScaleY}{\gdef\tsGraphicsScaleY{.9}}{}
\def\checkGraphicsWidth{\ifdim\Gin@nat@width>\linewidth
	\tsGraphicsScaleX\linewidth\else\Gin@nat@width\fi}

\def\checkGraphicsHeight{\ifdim\Gin@nat@height>.9\textheight
	\tsGraphicsScaleY\textheight\else\Gin@nat@height\fi}

\def\fixFloatSize#1{}%\@ifundefined{processdelayedfloats}{\setbox0=\hbox{\includegraphics{#1}}\ifnum\wd0<\columnwidth\relax\renewenvironment{figure*}{\begin{figure}}{\end{figure}}\fi}{}}
\let\ts@includegraphics\includegraphics

\def\inlinegraphic[#1]#2{{\edef\@tempa{#1}\edef\baseline@shift{\ifx\@tempa\@empty0\else#1\fi}\edef\tempZ{\the\numexpr(\numexpr(\baseline@shift*\f@size/100))}\protect\raisebox{\tempZ pt}{\ts@includegraphics{#2}}}}

\AtBeginDocument{\def\includegraphics{\@ifnextchar[{\ts@includegraphics}{\ts@includegraphics[width=\checkGraphicsWidth,height=\checkGraphicsHeight,keepaspectratio]}}}

\DeclareMathAlphabet{\mathpzc}{OT1}{pzc}{m}{it}

\def\URL#1#2{\@ifundefined{href}{#2}{\href{#1}{#2}}}

\def\UrlOrds{\do\*\do\-\do\~\do\'\do\"\do\-}%
\g@addto@macro{\UrlBreaks}{\UrlOrds}

\edef\fntEncoding{\f@encoding}

\makeatother

\newif\ifmultipleabstract\multipleabstractfalse%
\usepackage{tabulary}
\makeatletter
\AtBeginDocument{\@ifpackageloaded{longtable}{%
\def\LT@makecaption#1#2#3{%
  \LT@mcol\LT@cols c{\hbox to\z@{\hss\parbox[t]\LTcapwidth{%
    \sbox\@tempboxa{#1{#2: } #3}%
    \ifdim\wd\@tempboxa>\hsize
      #1{#2: }\textsc{#3}%
    \else
      \hbox to\hsize{\hfil\box\@tempboxa\hfil}%
    \fi
    \endgraf\vskip\baselineskip}%
  \hss}}}
}{}}
\makeatother

   \makeatletter
  \def\fig@textbf{\textbf}
   \AtBeginDocument{\renewcommand\floatc@plain[2]{\setbox\@tempboxa\hbox{{\footnotesize#1.}\footnotesize\hskip.5em#2}%
    \ifdim\wd\@tempboxa>\hsize {\fig@textbf{\footnotesize#1.}}\footnotesize\hskip.5em#2\par
        \else\hbox to\hsize{\hfil\box\@tempboxa\hfil}\fi}}
    \makeatother
  
\usepackage{float}

\begin{document}

%
% paper title
% Titles are generally capitalized except for words such as a, an, and, as,
% at, but, by, for, in, nor, of, on, or, the, to and up, which are usually
% not capitalized unless they are the first or last word of the title.
% Linebreaks \\ can be used within to get better formatting as desired.
% Do not put math or special symbols in the title.

% conference papers do not typically use \thanks and this command
% is locked out in conference mode. If really needed, such as for
% the acknowledgment of grants, issue a \IEEEoverridecommandlockouts
% after \documentclass

        \title{Persistent And Scalable JADE: A Cloud based In Memory Multi-agent Framework}
      \author{
		\IEEEauthorblockN{Nauman~Khalid}\\[-12pt]Email: 11mscsnkhalid@seecs.edu.pk 
        \vspace*{1pc}\and 
		\IEEEauthorblockN{Ghalib~Tahir}\\[-12pt]Email: 12mscsgtahir@seecs.edu.pk 
        \vspace*{1pc}\and 
		\IEEEauthorblockN{Peter~Bloodsworth}\\[-12pt]Email: peter.bloodsworth@cs.ox.ac.uk  ~\\(Corresponding author)}
  
% use for special paper notices
%\IEEEspecialpapernotice{(Invited Paper)}

% make the title area

\maketitle 
% Note that keywords are not normally used for peerreview papers.

% For peer review papers, you can put extra information on the cover
% page as needed:
% \ifCLASSOPTIONpeerreview
% \begin{center} \bfseries EDICS Category: 3-BBND \end{center}
% \fi
%
% For peerreview papers, this IEEEtran command inserts a page break and
% creates the second title. It will be ignored for other modes.
\IEEEpeerreviewmaketitle

\section{1. Abstract}
Multi-agent systems are often limited in terms of persistence and scalability. This issue is more prevalent for applications in which agent states changes frequently. This makes the existing methods less usable as they increase the agent's complexity and are less scalable. This research study has presented a novel in-memory agent persistence framework. Two prototypes have been implemented, one using the proposed solution and the other using an established agent persistency environment. Experimental results confirm that the proposed framework is more scalable than existing approaches whilst providing a similar level of persistency. These findings will help future real-time multiagent systems to become scalable and persistent in a dynamic cloud environment.
    
\begin{IEEEkeywords}multi-agent systems, cloud, jade, agents, Presistent, Scalable\end{IEEEkeywords}
\section{2. Introduction}
In this modern era Multi-agent systems are being used in many areas including aircraft maintenance, electronic book buying, network security, military logistic planning and maintaining adhoc networks. Organizations are transferring their large-scale systems to agent-based architectures like in telehealth care \unskip~\cite{821157:19647929}\unskip~\cite{821157:19647928}, fault diagnosis in the provision of an internet business\unskip~\cite{821157:19647927}  and many others\unskip~\cite{821157:19647934}\unskip~\cite{821157:19647926}\unskip~\cite{821157:19647925}.As Multi-agent system does not suffer with; resource limitations, critical failures or single point of failure, resource allocation deadlocks or bottlenecks and improves computational efficiency, reliability, flexibility, robustness and performance efficiency of the overall system\unskip~\cite{821157:19647933} . Although there are many advantages of multi-agent system there are several challenges distributed global environment. One of the vital issue which multi-agent system faces is a persistency \unskip~\cite{821157:19647924}\unskip~\cite{821157:19647923}\unskip~\cite{821157:19647922}\unskip~\cite{821157:19647921} of an agent because agents tend to run in memory and exists in a range of complex states. An environmental state of an agent is central to the autonomy of the system and therefore difficulties occur if an agent unexpectedly dies for some reason. When one dies all in-memory state information is lost. Especially in the scenario of those smartphones and computer applications in which agents are performing tasks on user's behalf and are communicating with each other. Such systems are mapping single user to personalized agent. When agent dies unexpectedly it results in a loss of important data of user and connection of a user with other users.

There are several approaches which are used by Java Persistence API\unskip~\cite{821157:19647915,821157:19647914,821157:19647913}, Serialization mechanism\unskip~\cite{821157:19647914} , DBMS\unskip~\cite{821157:19647913} and JADE Persistence Services\unskip~\cite{821157:19647911}. Java persistency framework did improve the flexibility and stability of an agent-based system however it increases the complexity of agents whose state changes in real-time. In the case when a certain object of agent is persisted in the database using composite keys, there is an increase in complexity to persist and find the object. Increase in artifact size, framework complexity, memory for a single agent results in frequent failures, especially in the scenario when the agent state is updated frequently. There is also a chance of Virtual Machine instance failure which will further increase the recovery time of the agent placed on a single instance due to VM churn time. 

The serialization mechanism is appropriate for short term storage of arbitrary objects. It is mostly used to persist the session information. However, when the data persistence is for a longer period or the state of the object frequently updates, the serialization cannot update the state of the agent to the session database in real-time. Moreover, when there is a change in any data structure of the agent, the serialized object cannot be deserialized even when you know the session id. 

Existing Jade persistence addon uses Java persistence API and relational DB and inherits their disadvantages for real-time multi-agent applications. The usability is further decreased in the cloud environment when persisting object state on an instance increasing the recovery time in case of VM failure. In order to address these issues we have purposed a novel approach. Our framework is based on real-time two-way communication in a similar manner to that of human interaction. Each agent has a mirror agent which duplicates its internal state. They communicate with each other in order to maintain up to date state information. If one of the agents should crash or die then the other agent takes its place and will recreate the agent that died. This makes the platform far more stable as mirror agents reside in a different physical location. We have evaluated our approach by comparing it with widely used Jade persistency framework\unskip~\cite{821157:19647911}.
    
\section{3. Background}
In Multi-agent system agents will have many states. Such states can be captured either in system level or application level. The capturing of state at the system level will result in a very generic and concrete solution but it is really tough and it would need some comprehensive coding in the Java virtual machine (in case of JADE) or agent development framework. Another way of capturing and persisting agent states is at the application level where the states of system are captured at run-time. Multi-agent systems have agents which collaboratively work to complete certain tasks which mean they share their some states with other agent in multi-agent system; this sharing creates collaboration between agents which can be said multi-agent environment. This sharing requires the classification of states with respect to responsibilities of agent.

Bracciali et al \unskip~\cite{821157:19647939}  proposed a model to classify states of agents of multi-agent systems. The model obtains the all possible states in multi-agent systems and broadly categorizes them in two sets, Environmental states and Agent states. Environmental states are states of agents which are shared with other agent in multi-agent systems while Agent states specifically belong to that agent itself only. The model for modeling the agent state should be generic so in this paper agent is consider as black box which means states are model regardless of working of agent.The actions of agents are being observed to capture states and these actions also change the states of agent. These actions of certain agents are observed by other agents who in response take a specific course of action. Each agent maintains its mental state containing its beliefs, desires and what it intends to do. \mbox{}\protect\newline Further, in the proposed model there are fully transparent Multi-agent system and partially transparent multi-agent systems. In transparent Multi-agent systems each agent knows the mental state of all other agents in system while in partial Multi-agent systems each agent has certain limited visibility of mental state of other agent \unskip~\cite{821157:19647936}. The work is significantly useful for classifying agent states. Although this process is very complicated, this increases the complexity of Multi-agent systems. Using this approach as ads on or frameworks will result in increasing complexity of systems. Similar approach with some modification can work notable well.

However, the classification of states of agents in Multi-agent systems reduces the complexity of overall systems but complexity of internal states of each agent should also be determined. The state of each individual agent can change frequently as the environment around them is continually evolving. If a database is used to store state information then it can become a serious bottleneck in terms of performance and scalability. This in turn reduces the scalability of the system and also makes it more prone to reliability issues. When an agent dies unexpectedly then the time required to recreate an Agent is critical. This is because agent will continue to receive action messages. There are several techniques which are presented as a solution of this problem but all of them have some shortcomings. Approaches which are currently being used in order to address this problem include: Java Persistence API, Serialization mechanism, DBMS and JADE Persistence Services. These persistence frameworks are discussed in detail below along with their shortcomings. 

The Java persistence API\unskip~\cite{821157:19647915} is a framework for developing business logic of enterprise application that edges the speed, security, and reliability of server-side technology. Caire et al \unskip~\cite{821157:19647938} developed agent based mobile application for hybrid networks which are combination of several networks including GPRS, UMTS and WLan. They implemented an application in J2ME and used JADE as the agent development framework. They intensively checked scalability and stability of the application. They deployed application on network of wireless devices. In order to check scalability, interoperability and fault tolerance, the information in MTP table, Container tables and Global Agent Descriptor Table are evaluated. In terms of persistency they preferred the Hibernate framework \unskip~\cite{821157:19647919} which provides object relational mapping (ORM). Hibernate has been proven to be able to successfully provide flexible and stable persistency because of its having support for the object query language.\unskip~\cite{821157:19647937} \mbox{}\protect\newline The Java persistency framework improved the flexibility and stability of an agent-based system. However, its scalability was not satisfactory for agents who frequently change their state and thereby send huge numbers of requests to the persistency framework. Hibernate has the capability to perform secure and structural transactions but its performance degrades as the burden on it increases. It was added as add-on in the JADE platform for providing application level persistency but the developers have described that there is still need to improve this add-on for highly scalable agent based systems \unskip~\cite{821157:19647918}. 

Serialization is another way being used for persisting agent's states in MAS. Serialization is a mechanism for representing an object in the form of bytes which contain information about the object and the type of data that is stored. Wong et al \unskip~\cite{821157:19647937}  designed a Java based infrastructure for the development of a network and mobile agent based application named Concordia. Concordia is middleware infrastructure which is capable of accessing information from wireless and wired devices that are connected to a network. This framework has many parts which allow agent administration, communication, migration etc.

Along with them, there is component for agentpersistence which allows the system to recover an agent's state when it dies. In this infrastructure they used Java object serialization \unskip~\cite{821157:19647940} . Serialization is usually preferred for persisting an object for short periods of time and in cases where object data does not change frequently because changes in class format of an object makes it hard to desterilize the Java object. In contrast agent states can change rapidly so serialization is not really appropriate for addressing persistency problems.

Database\unskip~\cite{821157:19647913} is widely used technique for persisting agent states. States can be stored in a database which then allows them to be fetched, updated and deleted. Databases provide a very structural way for these operations. If we consider the use of databases for persistency in a MAS there is a significant level of complexity. Each agent has its own internal state which may or may not be different from the other agents in the environment. In a database schema these states will be attributes of table and agent states possibly change on changing the responsibilities of agent. Such responsibilities will increase or decrease numbers of states which will eventually change the schema of database which results in broken keys and records. This will actually increase the complexity for database because  this cannot achieve using simple design. This complexity will increase with the passage of time and consequences in crashes and instability. Hence this change in responsibilities results in change of attributes of table and that continuous updating of attributes will make database inconsistence and generate errors. Another reason for not selecting databases as a persistency mechanism for multi-agent systems is their response times. Databases are used for storing data and their performance is hardware depended. Consider the case of a Multi-agent system which could have several thousands of agents distributed across a network. The agents in multi-agent system in some cases frequently change its states results in a massive amount of requests to the database. This response time of database can increase by enhancing hardware resources but it will be very costly.

JADE persistency services\unskip~\cite{821157:19647911} (JPS) has core module for providing persistency services known as JADE object manager. It is responsible for managing and persisting Java object using object oriented JADE database. The JPS are available as add-on of JADE which is not complete so it is not feasible for using in large applications. It does not support system level persistence yet, it is an add-on which eventually saves agent states in Database using an object relational model \unskip~\cite{821157:19647920}. So far it does not support system level persistence and also application level persistency is not fully implemented and need some concrete implementation of some modules. It does not provide sufficient support so as to be used in enterprise multi-agent systems. The feedback of persistence services are not satisfactory and review of different forum showed that using persistency services is not a better choice.

In above methods, agent persistency is provided but the approaches are not really at all efficient and work is therefore required to improve on scalability. These techniques had such shortcomings because of complex internal states of agents. For example in the Concordia infrastructure\unskip~\cite{821157:19647937}, serialization was used. Serialization is not ideal for large objects and does not perform well in terms of object access and transaction control (especially in the case of concurrent access.) Using a DBMS to store the states of agents persistently may limit the scalability of the system. In large-scale systems where agents rapidly update their states, they may need to frequently access a database server which can result in a server becoming overloaded.
    
\section{4. Multi-agent Persistency Framework}
The architecture of proposed approach included many agents. These agents and their states were classified in different groups. The agents were classified with respect to their responsibilities in multi-agent systems. The states of these agents were grouped in contrast of their scope. Scope was visibility of an agent states to other agent in multi-agent systems. Several agents worked together to run complete architecture of proposed approach. These agents were discussed below including their responsibility and location in architecture of proposed approach. The next section contains their working and how all of them collaborate to run the system. In proposed multi-agent architecture these agents are named as; working agents, clone agents, Local monitoring agents and Remote monitoring agents. This classification of states and agents actually defined the rules of overall system such as visibility rules for agent's states, placement of agents in framework and responsibilities of each agent.

In our proposed framework the internals of multi-agent systems were divided into environmental and local states. Environmental states involved information within a multi-agent system that was accessible by all agents. These states represented the current status of multi-agent system and might include blackboards and other data structures. Local states belong to an agent's internal structure. These states may or may not be visible to other agents in system. The majority of local states remained private and these states were not shared with other agents. In a few relatively rare cases some states might be shared with monitoring agents. Both states were saved in serializable and encapsulated objects which had methods through which they could be accessed and managed.

These objects were serialized so that they could be transferred as the content of ACL message. A serialized object held environmental state information and was associated with the Monitoring agent. Another group of objects was used to store the internal states of each agent.

Working agents in multi-agent systems had certain responsibilities that contributed in accomplishing overall objective of system. These were autonomous agents that performed their tasks independently and shared their information with other agent using communication. Their tasks were divided among agents depending upon requirements of system. Each agent in system could have different task or set of agents was assigned single task, so multi-agent systems had agents with different internal states. In this paper these agents were putted in one category although they had different internals. Because approach was proposed to restore these agents when they died so these were most crucial part of multi-agent systems.

The clone agents were mirror agents of their respective worker agents. Each worker agent had its clone which kept internal states of respective agent. Clone agent did not have its own internal states; it kept synchronized states of its working agent. The worker agent and its clone agent could be in same platform or in different platform. In proposed approach we resided clone agents in different platform so that each platform could send and receive request of other platform such as client server architecture. This was helpful to analyze performance of communication between clone agent and its worker agent. The in-depth scenario of their communication and re-creation of these agents including clone and worker agent was discussed later in this section.

Local monitoring agent controls and supervises worker agents. These agents may be one or more than one, depending on number of working agent in platform. These agents were responsible for creating died agents and managing working agents and allowed agents to communicate with agent in other platform. When agent died local monitoring agent was responsible to create agent with synchronized states. The synchronized states were provided by respective clone agent of died agent. These agents also managed the environmental states of multi-agent systems and other information including memory, CPU utilization, and platform information. Platform information included different addresses of platform, number of container and agents in these containers. Remote monitoring Agent resided on platform having clone agents. These agents created clone on request of worker agent. Remote Monitoring agent may be one or more than one depending on scalability of multi-agent systems. Like local monitoring agent it collects information from its platform and share with other platform.

Just logical divisions of states of multi-agent systems were not enough to manage such unstructured data. These states were associated with concern agents, therefore local states were assigned to its respective agent and environmental states were given to monitoring agents using an object encapsulation. These states were encapsulated in object which was serialize-able having methods to access and manage the states. The ACL message only allowed serialized object as content to transfer it between agents. Serialized object which contained either environmental or local state could be sent to any other agent in ACL message. Object having environmental states information was associated with monitoring agents and all agents can updates and change these states. On the other hand local states were not shared with even monitoring agent. These states were associated with agent and other group who had responsibility to manage them. Working agent shared these states to its clone agent which was used by them to recreate agent when the agent dies. Considering the case where clone died; the clone recreated using synchronized states of respective working agent.

Another challenge in proposed approach is to move agent from one platform to another. Although JADE gives support for moving agents from between platforms but does not transfer agent with its states information. An agent has code and states where code is actually behavior of agent and states are information of agent. These states show current internal state of agent and it uses these states for decision-making. Secure migration of agents is more challenging with code and states information. Migrating agents with code and states information could result in more content on a channel which will eventually result in latency and delay. The proposed approach for migrating agents from one platform to other is very effective because it is communication based and agents are good at learning from environment via communication. Also, FIPA specification contains interaction protocols, content language representation, blocking and unblocking communication support which improves the communication capabilities of agents. Ametller et al have presented a design and implementation which is capable of allowing agent migration over FIPA ACL messages\unskip~\cite{821157:19647917}. 

In the system there were a minimum of two JADE platforms; one was running on the local machine whilst the other was deployed in the cloud (in our prototype system this was the Amazon EC2 cloud\unskip~\cite{821157:19647910} .) The local monitoring agent and remote monitoring agents led local and remote platforms respectively. The number of monitoring agents could be scaled to match the size of platform in which they were operating. The overview of proposed approach is shown in Figure 1.

\bgroup
\fixFloatSize{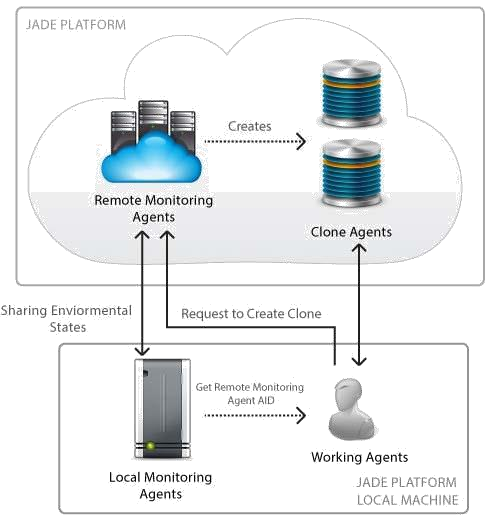}
\begin{figure}[!htbp]
\centering \makeatletter\IfFileExists{images/7e9fd5c5-f642-46bb-9493-676ca4f80be0-uarchitecture.png}{\includegraphics{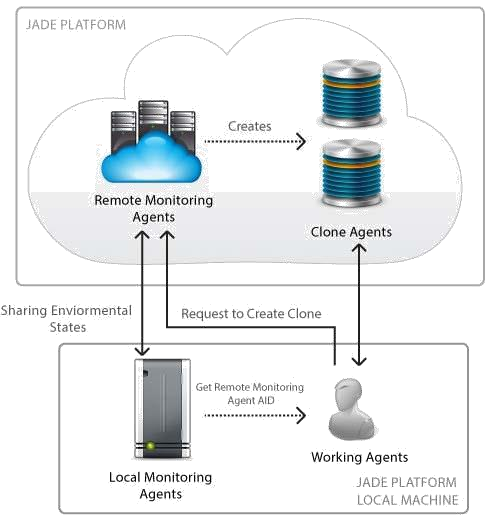}}{}
\makeatother 
\caption{{Architecture diagram Proposed Approach}}
\label{figure-9d36c2603d6b3c1e00b4e5a39ba462ae}
\end{figure}
\egroup
Initially both local and remote monitoring agents shared their AID's with each other. The AID is a unique agent identifier containing the agent name and its addresses. Agents running in the local machine got a remote monitoring agent AID from the local monitoring agent and sent requests for clone agent creation (clones act as mirrors of their respective working agents.) This is carried out via an ACL message which contained serialized local state objects. The remote monitoring agent created the requested clone with the initial states that were provided. The clone and working agent then began direct communication with each other. The working agent sent a stream of state updates to its clone. This is a similar mechanism that humans might use in an organization when they cc a colleague into an e-mail thread so that they were aware of what is happening and can take over if the need arises.
    
\section{5. Evaluation And Results}
In order to evaluate the performance of the proposed approach, we have compared it with state-of-the-art Jade persistence framework. This framework was found to be the latest approach and considered to be the most stable and efficient. Also, it was available as plugin in JADE platform with good support which made implementation less challenging. These prototypes were implemented with same functionality and techniques to ensure a fair comparison.

An agent-based auction simulation system was selected as a good use-case for evaluating the system. This was because it is a real-time problem which has quite a high level of communication and in which internal agent states were likely to be frequently updated. Using this prototype we evaluated the core features of our system in a real- world Multi-agent environment. We wanted to compare our framework against the state-of-the-art system. We therefore built the auction simulation system using both our approach and the well-known JADE persistency framework. Both systems were then evaluated qualitatively and quantitatively. The qualitative evaluation included the testing of prototype functionality and quantitative evaluation was measurement of performance. Quantitative evaluation was performing using a number of metrics including: response time, CPU and memory usage and agent recreation time.

\subsection{5.1. Experimental Setup}In this section we have done the comparative study of prototypes of simulating auction system. The auction modules of prototypes were deployed on local machine whilst persisting modules were placed on an Amazon EC2 micro instance. The persisting module in proposed technique was a Multi-agent system. This system was developed using the JADE framework. A platform deployed on local machine was called the local platform while a platform which was deployed on Amazon Ec2 instance was called Remote Monitoring platform. Each platform had only monitoring agent known as local monitoring agent and remote monitoring agent. The working agents were placed in local platform and clone agents were on remote platform. These two platforms were running independent of each other and they communicated using ACL messages. Each working agent created by a system maintained its local state while the monitoring objects managed the environmental states.

MTP URL was used to access remote platform and agent name helped to deliver a message to particular agent. The working agent then sends a message to create a clone agent. After the clone agent was created it sends the acknowledgement message to its working agent. Then after each bid, working agent sent its latest local states to its clone agent. This way clones stored the local states of their respective working agents.

Prototype developed using JPF has a platform which ran on a local machine. It has monitoring agent named as Main agent. This agent kept track of all working agent and environmental states. It manages the auction system by maintaining data structure. It had list of all working agent and notify agents to bid in queue. The working agents register themselves with main agent by sending their states. The system is capable of persisting states of agent and recreating agent when dies. Whenever any agent would die, main agent will fetch it states from database and recreate it with updating states. The detailed flow of prototype is shown in Figure 2.

\bgroup
\fixFloatSize{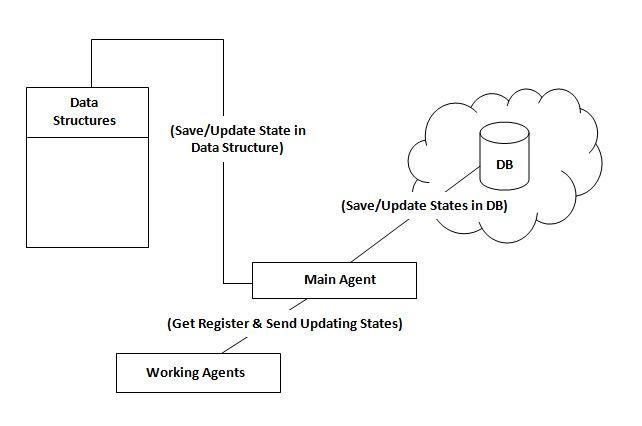}
\begin{figure}[!htbp]
\centering \makeatletter\IfFileExists{images/a8cd040d-9c15-4e9b-a14e-7c566ba90ba5-uarchitecturejpa.png}{\includegraphics{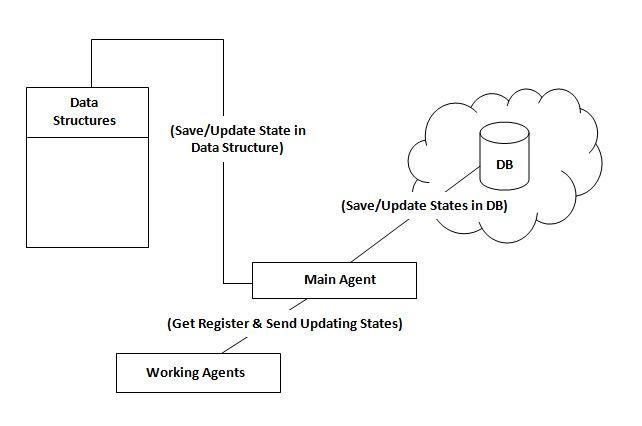}}{}
\makeatother 
\caption{{Architecture of JPA}}
\label{figure-fa251fd7d673f7f0f71ac8c7b38baae6}
\end{figure}
\egroup
Before quantitative evaluation, the prototypes were deployed and tested qualitatively. This testing included the functionality testing of different modules and features by running several test cases. This testing ensured that prototype were ready for evaluation process. In order to evaluate the performance of the prototype certain parameters have been identified. These parameters include memory usage, CPU usage and response time of the system. The detailed hardware specifications are listed below.

The local machine used for deploying and evaluating prototypes was HP 630 laptop. The technical specification of laptop is listed in Table 1.

\begin{table}[!htbp]
\caption{{ Specification of Local Machine} }
\label{table-wrap-08a79d4dae16f24744560bf536e94b9b}
\def\arraystretch{1}
\ignorespaces 
\centering 
\begin{tabulary}{\linewidth}{LL}
\hline 
Processor &
  Intel Core i5-2430M\\
RAM &
  2GB DDR3\\
Operating System &
  Window 8, 32bit\\
\hline 
\end{tabulary}\par 
\end{table}
The persistence units of both prototypes were deployed on T2 micro instance. This is a general purpose instance, which is suitable for workloads such as web server or IDE. The specifications of the T2 micro instance, are listed in the Table 2.

\begin{table}[!htbp]
\caption{{ Specifications of Amazon Instance} }
\label{table-wrap-c634dc70424d39e7704ca684cbca305d}
\def\arraystretch{1}
\ignorespaces 
\centering 
\begin{tabulary}{\linewidth}{LL}
\hline 
Processor &
  Intel Xeon Processors\\
Processor speed &
  2.5GHz to 3.3 GHz\\
CPU Credit/ hour &
  6\\
RAM &
  1 GB\\
Operating System &
  Linux\\
Storage &
  EBS\\
\hline 
\end{tabulary}\par 
\end{table}
An experiment was setup to get results of both types of evaluations. The qualitative results were obtained using constant number of agents, they were ten and for quantitative evaluation the number of agents was increased to measure scalability of framework. Response time is the length of time taken by an agent to register with the monitoring agent, save its states and become part of the auction system. CPU usage and memory usage are the overall levels that were consumed by the systems. The time between an agent dying and its recreation is called the agent recreation time. A gradually increasing set of tests were carried out with complexity being added by expanding the number of bidding agents in the system.

\subsection{5.2. Qualitative Results}Qualitative results mainly concern with functionality and quality of the prototypes. This testing was performed to test working of core modules of both persistency frameworks. Although the prototypes were simulating auction system but using different persistency approaches. It was important to make sure that key features of prototypes were working. We have got qualitative results to ensure that both prototypes are performing same functionalities. The prototypes can fairly compare if and only if they are doing same functionalities.

 The qualitative results of proposed approach were gotten comprehensively. There were several features of prototypes and most of them are similar. We have implemented both prototypes using same logic but steps involved in persistency frameworks were different. The purpose of prototype developing using proposed approach was to store agent states and recreate them with latest states when they crash. The testing of key functions of proposed approach were; launching both platform and sync monitoring agents, adding bidders on user input, inter-platform communication, clone creation, working agents states synchronization with its clone agent, working agents recreation with latest states when single or several agents crash, Clone agents recreation when single or several clone agents crash and recreating monitoring agents. Each test case is executed several times for accuracy. In case of any unexpected results or failure in any test case, it was debugged and particular reason of this failure was identified and removed.

The qualitative results of prototype developed using Jade persistence framework are presented in order to analyze the working of its modules. The core objectives of both prototypes are similar but uses different approaches which consequences in different test cases. The number of test cases could be different for each prototype, because procedure of persisting agent states and recreating crashed agent are different. The key functionality of this prototype are launching JADE platform with working agents locally, adding bidders on user input, communication with Main Agent, agents connectivity with Database, and working Agents recreation with updating states when single or several agents crash.

\subsection{5.3. Quanitative Results}The response time of systems includes the time period, an agent took to; register itself with monitoring agent, persisting its state and take part in auction system. Considering the proposed approach, the response time of agent include; agent registration with monitoring agent, requesting remote platform for its clone and working agent synchronization with its clone agent. Similarly response time in prototype with Jade persistence framework is time it took to register with main agent, store its states in Database and ready to take part in auction system.

The graph in Figure 3 shows the database powered JADE persistency framework has a much slower response time than our system. This is because our system depends on agent communication and networks are generally pretty fast. Databases are much slower in terms of reading and writing which accounts for the big differences in performance between the two systems. As the numbers of agents were increased in our system the response time only increased slightly. This is because each agent in the system has its clone on the cloud. The agent and its clone directly communicate with each other thus minimizing the response time. For similar reasons the agent recreation time is reduced.

\bgroup
\fixFloatSize{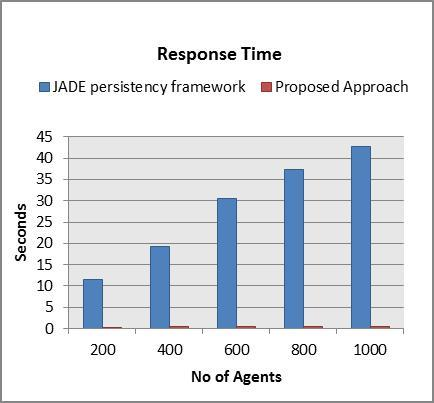}
\begin{figure}[!htbp]
\centering \makeatletter\IfFileExists{images/79fe261e-8529-459a-b17b-65467042c30c-ucomparison-of-response-time.png}{\includegraphics{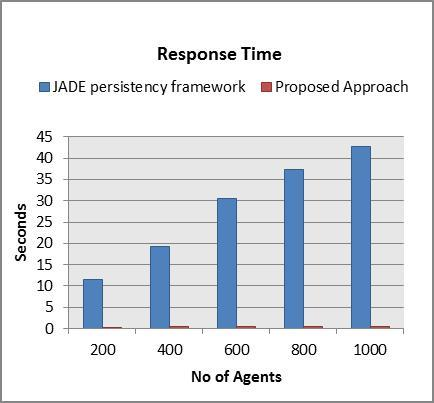}}{}
\makeatother 
\caption{{Comparison of Response Time}}
\label{figure-a44d92cecea2f815cf7cc9542deca0e8}
\end{figure}
\egroup
The response time of the proposed approach remains below 0.6 second for a thousand users. In contrast the response time of the JADE persistence framework increased to forty five seconds. The growth rate of response time of the proposed approach is shown in Figure 4 to clearly show the patterns in response time. The increase in response time is linear but it is a very slight increase. This slight increase in response time shows the improvement in the performance of the proposed approach. The approach is therefore considered as a good choice for systems where agents' states frequently change.

\bgroup
\fixFloatSize{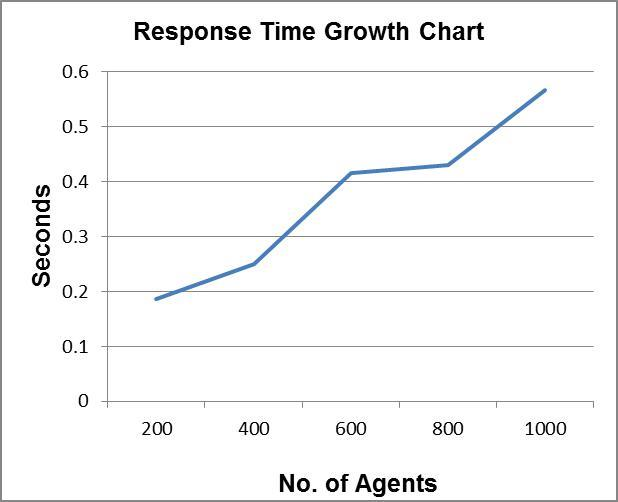}
\begin{figure}[!htbp]
\centering \makeatletter\IfFileExists{images/9611be95-4d9f-49e6-b232-263ebe0fb3cd-uresponsetimegrowth.png}{\includegraphics{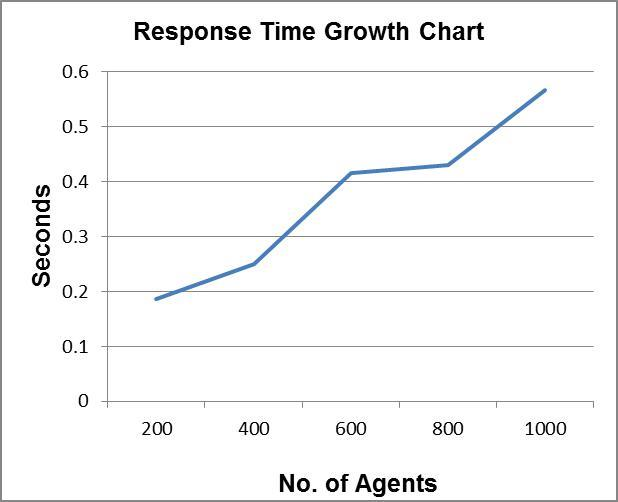}}{}
\makeatother 
\caption{{ Response Time Growth}}
\label{figure-74e917574c5838b5a9f7ee07cb8da5c6}
\end{figure}
\egroup

\subsection{5.4. CPU and Memory Usage:}Agent recreation time is one of the most vital features of prototype. The agents can crash while performing their tasks. Both prototypes are able to recover them with synchronized states. It is important to analyses the performance of our system with JPF. To compare the agent recreation time of the prototypes, we had crashed agents during their execution. Different numbers of agents were killed to analyze the performance of prototypes and evaluated net effect on the end results. For example the bidder wins the auction should remain same either agents were killed or not. The killed agents should be recreated by monitoring agent with synchronized states so that it can adjust in the environment.

Figure 5 shows a graph which compares the lengths of time both systems took to recreate agents. Several agents were killed in each system in order to gather results regarding this. It was established that the recreation time of both systems does not change rapidly as the number of agents increase. Our approach was however found to be much quicker in recreating agents. This is important because it reduces the risk of failures cascading throughout the system and means that it can recover quickly from the errors that occur. All of this makes our approach more stable and performant than the original database driven framework.

\bgroup
\fixFloatSize{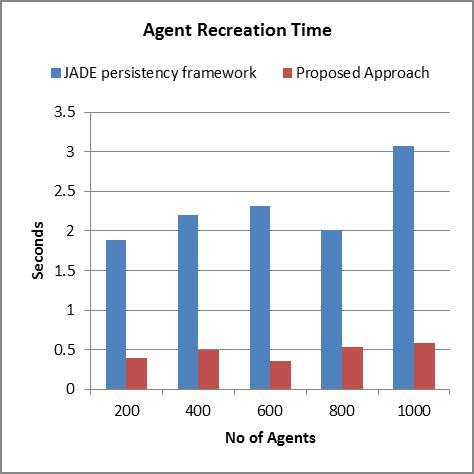}
\begin{figure}[!htbp]
\centering \makeatletter\IfFileExists{images/91f271d6-e725-4402-80cc-b91822b5c7cb-uagentrecreationtime.png}{\includegraphics{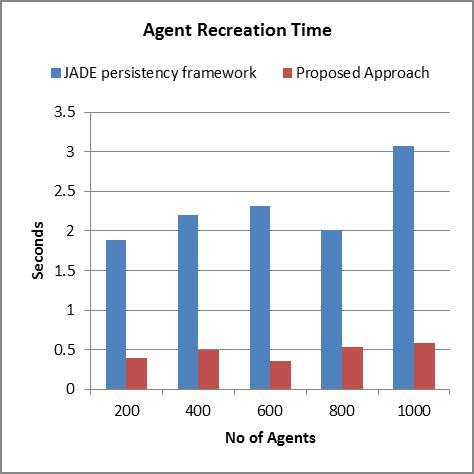}}{}
\makeatother 
\caption{{Comparison of agent recreation time}}
\label{figure-fa7d4fa083495ea5bf7737553d2b3744}
\end{figure}
\egroup
The overall CPU and memory analysis showed that proposed approach considerably efficient in CPU performance. Their usage patterns are used to compare the performance analysis of both prototypes. Their analysis will help us to understand the performance of both prototypes. The prototype is only considered to be useful when it can execute and deploy on reasonable computational resources. If prototypes demand for more computational and hardware resources, this would reduce the utilitarian aspects of the system. The new persistence approach should be resource effective as well. The Multi-agent systems can vary in number of agents; they can contain few agents or one enormous. The Multi-agent systems containing fewer agents would prefer to use less hardware resources. Similarly, multi-agent systems having enormous number of agents need resource effective approach to run smoothly.

Our system uses more memory than the database driven framework as can be observed in Figure 6. This is because the local monitoring agent's clones reside on the local machine and the agents running on it also consume memory. Our framework uses more memory as compared to JPF because in our solution we are persisting states in memory and are not that limiting in terms of scalability. This is the case because machines often have large amounts of available memory especially in the cloud where instances are available with many gigabytes of ram. This shows that the difference in memory usage is relatively small even when 1000 agents are running.

\bgroup
\fixFloatSize{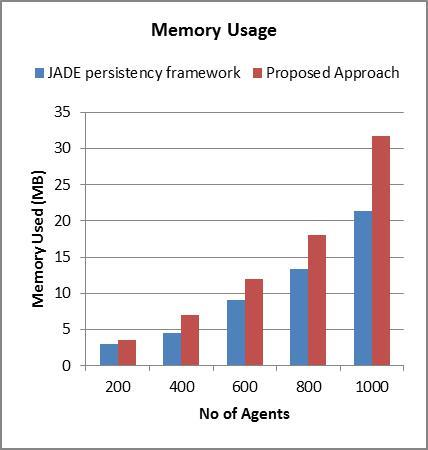}
\begin{figure}[!htbp]
\centering \makeatletter\IfFileExists{images/9d1765a0-7ebd-4599-8223-a636f9f7844c-umemoryusage.png}{\includegraphics{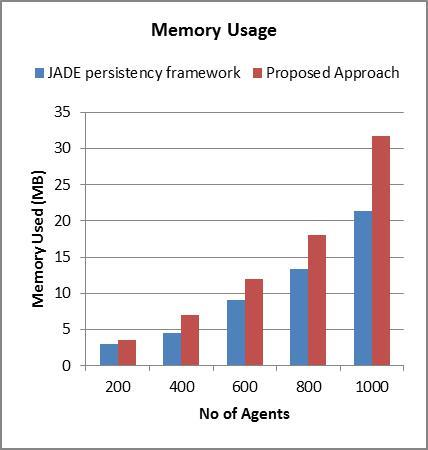}}{}
\makeatother 
\caption{{Comparison of Memory Usage}}
\label{figure-95c4240e4e807a5936cf3b86ea16dc6a}
\end{figure}
\egroup
The system launch time for the JADE persistency framework based auction system (including database initialization and agent registration) is around four minutes. On increasing the number of agents the system quickly crashes. The launch time for our system was around two and half minutes. This can however be improved by increasing the number of remote monitoring agents and distributing tasks between them more efficiently. In each test case the final result of the auction process was the same for both systems. This confirms that the systems are able to persistently manage the internal states of agents when they die. It also confirms that our approach provides a similar level of persistency to the original framework. \mbox{}\protect\newline Overall it can be seen that our approach significantly outperforms the previous state of the art persistency framework. Whilst some might argue that a database driven approach is more robust or permanent, the associated overheads and their impact on scalability and run-time error handling seems to make an in memory cloud based approach quite appealing. Making use of cloud resources can significantly enhance the scalability of the framework.
    
\section{6. Conclusion And Future Work}
In this paper we have proposed a novel approach of persistency for multi-agent systems. Our approach involves in memory mirroring of the internal states of agents using cloned agents in the cloud. Communication between regular agents and their cloud based clones used to synchronize the internal states of both agents. When either agent or its clone dies then the remaining agent takes over the role in the system whilst the agent is recreated. After an agent has been recreated a hand over process ensures that it has the correct internal state.

We have thoroughly evaluated our framework against the state-of-the-art JADE persistency service. This was carried out using a simulated auction as a use-case during experimentation. The results confirm the general performance benefits of our approach. Such benefits will enable our framework to scale better in complex environments. In the future we will explore the development of a procedure which can automatically adapt legacy MAS systems to make use of our proposed framework.

\section{\textbf{Funding}}
We like to acknowledge AMAZON for their kind support for this project.

\section{\textbf{Conflict of Interest}}
The authors declare that they have no conflict of interest.

% trigger a \newpage just before the given reference
% number - used to balance the columns on the last page
% adjust value as needed - may need to be readjusted if
% the document is modified later
%\IEEEtriggeratref{8}
% The "triggered" command can be changed if desired:
%\IEEEtriggercmd{\enlargethispage{-5in}}

% references section

% can use a bibliography generated by BibTeX as a .bbl file
% BibTeX documentation can be easily obtained at:
% http://www.ctan.org/tex-archive/biblio/bibtex/contrib/doc/
% The IEEEtran BibTeX style support page is at:
% http://www.michaelshell.org/tex/ieeetran/bibtex/
%\bibliographystyle{IEEEtran}
% argument is your BibTeX string definitions and bibliography database(s)
%\bibliography{IEEEabrv,../bib/paper}
%
% <OR> manually copy in the resultant .bbl file
% set second argument of \begin to the number of references
% (used to reserve space for the reference number labels box)
% \begin{thebibliography}{1}

\bibliographystyle{IEEEtran}

\bibliography{article.bib}
% \bibliography{\jobname}

% \EOD
\vfill
\end{document}